\documentclass[10pt,twocolumn,letterpaper]{article}

\usepackage[pagenumbers]{cvpr} 

\definecolor{cvprblue}{rgb}{0.21,0.49,0.74}
\usepackage[pagebackref,breaklinks,colorlinks,allcolors=cvprblue]{hyperref}

\usepackage{makecell}
\usepackage{tabularx}
\usepackage{multirow}
\usepackage{tikz}
\usepackage{caption}
\usepackage{graphicx}
\usepackage{subcaption}
\usepackage{float}

\def\confName{CVPR}
\def\confYear{2025}
\usepackage[normalem]{ulem}

\title{Neural Motion Simulator\\ Pushing the Limit of World Models in Reinforcement Learning}

\usepackage[switch]{lineno} 

\author{
  Chenjie Hao$^{1\ast}$, 
  Weyl Lu$^{1\ast}$, 
  Yifan Xu$^{2}$, 
  Yubei Chen$^{1,2\dagger}$\\[0.3em]
  \normalsize
  $^1$UC Davis \quad
  $^2$Open Path AI Foundation \\
  \tt\small \{cjhao, adslu, ybchen\}@ucdavis.edu, yifan.xu@openpathai.com
}

\begin{document}
\twocolumn[{%
\renewcommand\twocolumn[1][]{#1}%
\maketitle

\vspace{-1em}
\includegraphics[width=\linewidth]{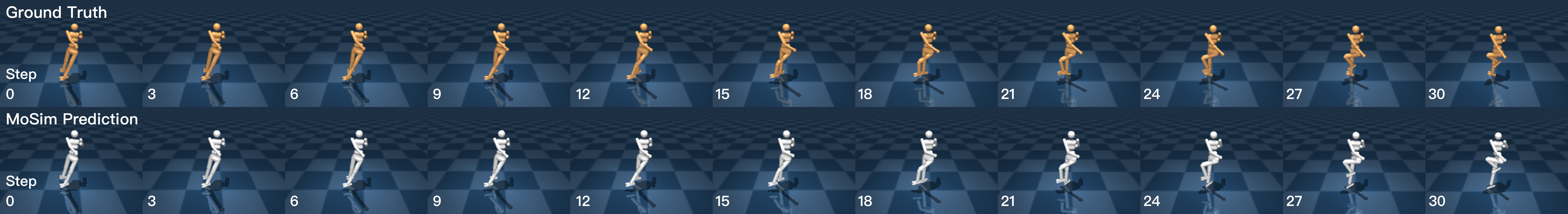}
\includegraphics[width=\linewidth]{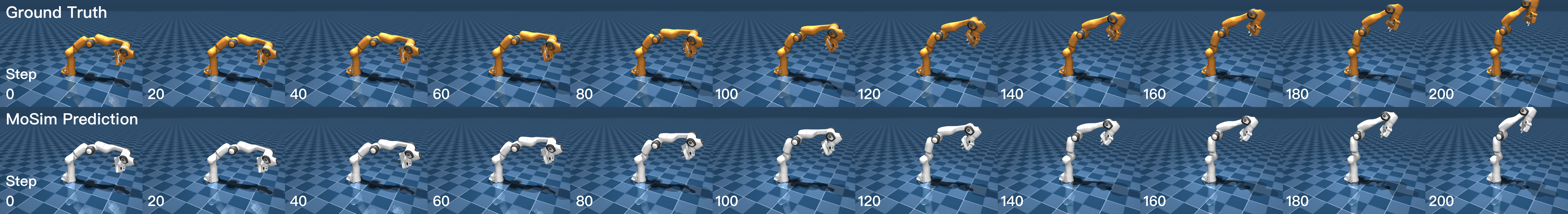}
\includegraphics[width=\linewidth]{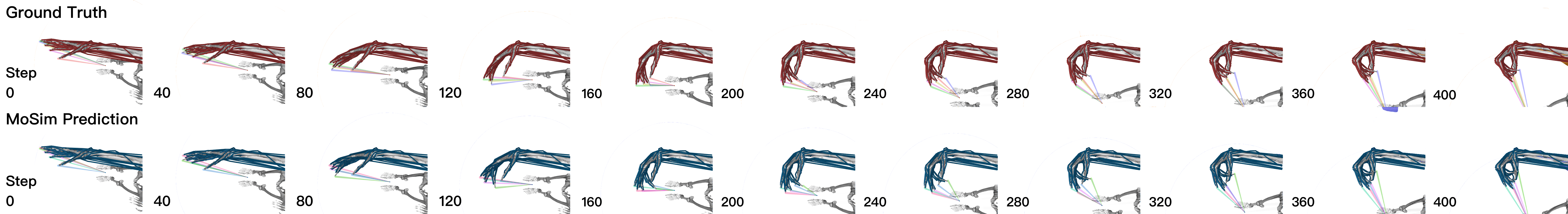}

\vspace{-0.5em}
\captionsetup{hypcap=false}
\captionof{figure}{This figure demonstrate the long-horizon precise prediction by the Neural Motion Simulators. In each of the three pictures, the first row shows the ground-truth states and the second row shows the predicted states with the same initial condition and actions sequence. Humanoid predicts for 30 steps with rendering every 3 steps; Panda predicts for 200 steps with rendering every 20 steps; myohand predicts for 400 steps with rendering every 40 steps. \vspace{1.5em}}
\captionsetup{hypcap=true}\label{fig:teaser}}]

{\let\thefootnote\relax\footnote{$^*$Equal Contribution. $^\dagger$Corresponding Author}}

\begin{abstract}
An embodied system must not only model the patterns of the external world but also understand its own motion dynamics. A motion dynamic model is essential for efficient skill acquisition and effective planning. In this work, we introduce the neural motion simulator (MoSim), a world model that predicts the future physical state of an embodied system based on current observations and actions. MoSim achieves state-of-the-art performance in physical state prediction and provides competitive performance across a range of downstream tasks. This works shows that when a world model is accurate enough and performs precise long-horizon predictions, it can facilitate efficient skill acquisition in imagined worlds and even enable zero-shot reinforcement learning. Furthermore, MoSim can transform any model-free reinforcement learning (RL) algorithm into a model-based approach, effectively decoupling physical environment modeling from RL algorithm development. This separation allows for independent advancements in RL algorithms and world modeling, significantly improving sample efficiency and enhancing generalization capabilities. Our findings highlight that world models for motion dynamics is a promising direction for developing more versatile and capable embodied systems.
\end{abstract}    
\section{Introduction}
\label{sec:intro} 
Human and other animals build world models of the external world and themselves to efficiently learn skills, reason, and plan \cite{lecun2022path}. By using such a world model, natural intelligence can learn tasks with a small exposure or generalize to the situations they have never encountered. A world model \cite{sutton1991dyna, henaff2017model, ha2018world} is an inner model that simulates how the world evolves, and in essence it does one thing: Given the current state and action, it predicts the future state \cite{sutton1991dyna}. In this work, we focus on building the world model for motion dynamics. Given an arbitrary robot, a world model must learn the dynamics of the physical body and predict the future physical state given its current state and actions. While many pioneering works were proposed in the past \cite{hansen2022temporaldifferencelearningmodel, hansen2024tdmpc2scalablerobustworld, hafner2019dream, hafner2020mastering, hafner2023mastering} to solve different reinforcement learning tasks, almost all of these works evaluate world models indirectly using the downstream tasks as surrogates. This work aims to fill this neglected gap. How good are the world models at doing what they do -- to predict the future given the current state and action? If the world models are truly generalizable, can we possibly use them to generalize to the never encountered situations or even learn the new tasks in a zero-shot manner?

\noindent In this work, we push the limit of world models in reinforcement learning and build the state-of-the-art world model, {\it Neural Motion Simulator} (MoSim), that significantly surpasses the previous ones in direct evaluations, predicting the future raw and latent states. We show, for the first time, that when a world model's prediction horizon and accuracy are sufficient, it is possible to train or search for a new policy completely in the predicted space in a zero-shot manner. In order to model the motion precisely, we introduce a world model architecture with rigid-body dynamics and Neural Ordinary Differential Equations (Neural ODE) \cite{chen2018neural}, which enables accurate long-horizon predictions of future physical states. Since MoSim predicts in the raw state space, it can be combined with essentially any model-free reinforcement learning (RL) algorithm and turn into a model-based approach. This effectively decouples the modeling of physical environment from the development of the RL algorithm. Such a separation allows MoSim to benefit from the independent advancements from both world models and RL algorithms, significantly improving the sample efficiency and enhancing generalization capabilities. Our findings highlight that modeling world models for motion dynamics is a promising direction for developing more versatile and capable embodied systems.

\noindent\textbf{Our main contributions are summarized as follows:}%
\begin{itemize}[leftmargin=1.5em, topsep=0pt, itemsep=0pt]
    \item We propose \textbf{MoSim}, a neural simulator with \textbf{state-of-the-art} long-horizon prediction accuracy.
   \item Leveraging MoSim’s predictive capability, we show that it is possible to achieve \textbf{zero-shot model-based RL}, which can be integrated with any model-free algorithm.
   \item We take initial steps toward addressing the key challenges of zero-shot RL, demonstrating the potential of this direction.

\end{itemize}
We created a website \url{https://oamics.github.io/mosim_page/} to provide additional details, including a link to our open source code repository.
\section{Method}
\label{sec:method}

\begin{figure*}[htbp]
    \centering
    \includegraphics[width=\textwidth]{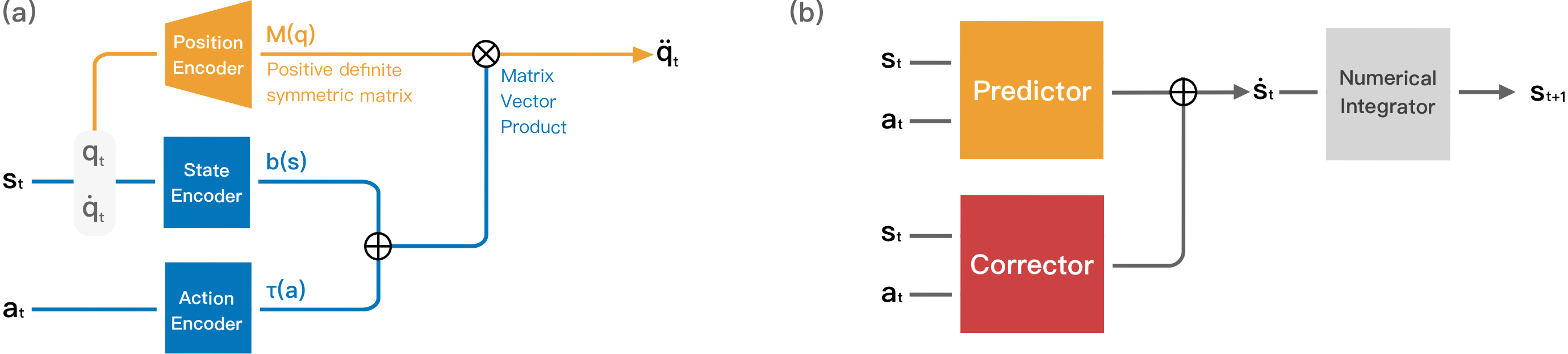}
    \caption{(a) Structure of predictor for a sub-step. (b) Structure of MoSim using Neural ODE to integrate many sub-steps to make a prediction for the next state.}
    \label{fig:model_structure}
\end{figure*}

\noindent In this section, we introduce the model architecture, benchmark, and training strategy.

\subsection{Architecture}
\label{subsec:arch}

\noindent {\bf Dynamics Model.} MoSim performs predictions within the continuous physical state space \( \mathcal{S}\) of the robot. The physical state \(\boldsymbol{s}(t) := (q_1,q_2,\dots,q_n, \dot{q}_1,\dot{q}_2,\cdots,\dot{q}_n)^T = (\boldsymbol{q}^T, \boldsymbol{\dot{q}}^T)^T\) is a vector that changes continuously over time $t$, including the coordinates $\boldsymbol{q}$ and their corresponding velocities $\dot{\boldsymbol{q}}$\footnote{In this paper, we use dots to denote derivatives with respect to time. For example, $\ddot{\boldsymbol{q}}$ represents the second-order derivative of $q$ with respect to time, i.e., acceleration.}. $\boldsymbol{s}$ includes variables such as joint motor angles, angular velocities, spatial positions and velocities, etc. In practice, these state variables can be obtained through sensors mounted on the robot. As shown in Figure~\ref{fig:model_structure}(b), We model the system dynamics by two terms: 1) a {\it Predictor} term $f$, which models the deterministic rigid body dynamics and 2) a {\it Corrector} term, which handles noise or other unmodeled factors that deviate from the rigid body motion: 
\begin{align}
    \dot{\boldsymbol{s}}(t) = \boldsymbol{f}(\boldsymbol{s}(t),\boldsymbol{a}(t)) + \boldsymbol{\epsilon}(\boldsymbol{s}(t),\boldsymbol{a}(t)),
\label{eq:dynamical_equation}
\end{align}
where \(\boldsymbol{a}(t)\) denotes a time-varying action vector, which can be joint torque inputs or other control signal inputs, \(\dot{\boldsymbol{s}}(t)\) represents the time derivative of \(\boldsymbol{s}(t)\), \(\boldsymbol{f}\) denotes the rigid body dynamic function, and \(\epsilon\) represents the residual unmodeled factors. In the absence of stochasticity, the initial state \(\boldsymbol{s}(t_0)\) and a given action \(\boldsymbol{a}(t)\) uniquely determine the state \(\boldsymbol{s}(t_0 + T)\) for any $T$. 

\noindent {\bf Ideal Rigid Body Dynamics.} For ideal rigid body motion, the dynamic equations have an explicit form \cite{goldstein2001classical,todorov2012mujoco}: 
\begin{align}
    \dot{\boldsymbol{s}}_{ideal} &= \frac{\mathrm d}{\mathrm d t} \begin{pmatrix} \boldsymbol{q} \\ \dot{\boldsymbol{q}} \end{pmatrix}  = \boldsymbol{f}(\boldsymbol{s}(t), \boldsymbol{a}(t)) \\
    & =
    \begin{pmatrix} 
    \dot{\boldsymbol{q}} \\ 
    M(\boldsymbol{q)}[\mathbf{b}(\boldsymbol{q},\dot{\boldsymbol{q}})+\boldsymbol{\tau}(\boldsymbol{a})+ \boldsymbol{c}(\boldsymbol{q},\dot{\boldsymbol{q}},\boldsymbol{a})]
    \end{pmatrix}.
\label{eq:rigid_body_equaitons}
\end{align}
where \(M(\boldsymbol{q})\) is the inverse of inertia matrix that depends only on the position and it is symmetric positive definite; \(\boldsymbol{b} (\boldsymbol{q},\dot{\boldsymbol{q}})\) is a vector-valued function of the state, describing the effects of conservative forces, such as gravity; \(\boldsymbol{c}(\boldsymbol{q},\dot{\boldsymbol{q}},\boldsymbol{a})\), as a residual term, accounts for the effects of equality constraints and contact forces (e.g., collision and friction forces) which are difficult to be explicitly modeled. 

\noindent If we let $\boldsymbol{\epsilon}(\boldsymbol{q},\dot{\boldsymbol{q}},\boldsymbol{a})$ in Equation~\ref{eq:dynamical_equation} absorb $M(\boldsymbol{q})\boldsymbol{c}(\boldsymbol{q},\dot{\boldsymbol{q}},\boldsymbol{a})$ in Equation~\ref{eq:rigid_body_equaitons}, then we can reach the following formulation:
\begin{align}
    \dot{\boldsymbol{s}}(t) = \begin{pmatrix} 
    \dot{\boldsymbol{q}} \\ 
    M(\boldsymbol{s})[\boldsymbol{b} (\boldsymbol{s}) + \boldsymbol{\tau}(\boldsymbol{a})]
    \end{pmatrix} + \boldsymbol{\epsilon}(\boldsymbol{s}(t),\boldsymbol{a}(t))
\label{eq:dynamic_final}
\end{align}

In MoSim, we parameterize each of $M(\boldsymbol{s})$ (Position Encoder), $\boldsymbol{b}$ (State Encoder), $\boldsymbol{\tau}$ (Action Encoder), and $\boldsymbol{\epsilon}$ (Corrector) using a different neural network, as shown in Figure~\ref{fig:model_structure}. We use standard ResNet \cite{he2016deep} to implement \(M\), \(\boldsymbol{b}\), $\epsilon$ and a simple MLP to implement $\tau$. Detailed configuration of these networks are left to the Appendix. We refer to \(f(\boldsymbol{s}, \boldsymbol{a}) =M(\boldsymbol{s})[\boldsymbol{b} (\boldsymbol{s}) + \boldsymbol{\tau}(\boldsymbol{a})]\) as the {\it predictor} and \(\epsilon(\boldsymbol{s},\boldsymbol{a})\) as the {\it corrector}. {\bf Predictor}. In the position encoder, the ResNet output is rearranged into a lower triangular matrix \(L\), from which \(M\) is calculated as \(M = LL^T\), based on the property of Cholesky decomposition \cite{golub2013matrix} that any symmetric positive definite matrix can always be expressed in this form. We  assembled $M$, $b$ and $\tau$ into $\ddot{\boldsymbol{q}}$, as illustrated in the Figure \ref{fig:model_structure}. Subsequently, $\ddot{\boldsymbol{q}}$ is combined with $\dot{\boldsymbol{q}}$ to form the output $\dot{\boldsymbol{s}}$. We embed strong inductive biases into the predictor by leveraging the general structure of rigid body dynamics. As demonstrated in the ablation study in Section \ref{sec:results}, these inductive biases significantly improve the prediction accuracy. It is important to emphasize that we do not incorporate any additional prior knowledge of physics beyond the compositional structure, and, in practice, our network does not rely on the physical interpretation of these intermediate variables. {\bf Correctors.} On the other hand, the correctors are made up entirely of one or more parallelly connected standard residual networks. Unlike the predictor, we do not introduce as much prior knowledge into the corrector, as its role is to handle the remaining complexities, such as friction, collisions, and unmodeled factors.

\noindent {\bf Neural ODE.} The general form of Neural ODE is a continuous extension of recurrent neural networks:
\begin{align}
    \frac{\mathrm{d} z(t)}{\mathrm{d}t} = g_\theta(z(t), t)
\label{eq:z_diff}
\end{align}
Here, \(z(t)\) represents the hidden variables over continuous \(t\), and the dynamical function $g_\theta$ is parameterized by a neural network. Its solution takes an integral form:
\begin{align}
    z(t_1) = z(t_0) + \int_{t_0}^{t_1} g_\theta(z(t), t) \, \mathrm{d}t
\label{eq:z_int}
\end{align}
The integration in practice is computed iteratively through numerical integration methods such as the Euler method\cite{euler1768integralis} and the Runge-Kutta method\cite{butcher2003runge, kutta1901runge}.  Our work employs the DOPRI5 integrator\cite{dormand1980runge}, which is an adaptive-order method based on the Runge-Kutta framework.

Neural ODE addresses the challenge of backpropagation in the continuous setting\cite{chen2018neural}, which is inherently more difficult than in the discrete case, by cleverly reformulating it as another numerical integration:
\begin{align}
    \frac{\mathrm{d}L}{\mathrm{d}\theta} = -\int_{t_1}^{t_0} \alpha(t) \frac{\partial g_\theta(z(t), t)}{\partial \theta} \, \mathrm{d}t , 
\label{eq:loss_int} 
\end{align}
where $L$ represents the loss function, and $\alpha(t):=\frac{\partial L}{\partial z(t)}$ is deined as the adjoint state of $z(t)$. The value of $\alpha(t)$ can also be obtained through an integral process.
\begin{align}
    \alpha(t_0) = \alpha(t_1) - \int_{t_1}^{t_0} \alpha(t) \frac{\partial g_\theta(z(t), t)}{\partial z(t)} \, \mathrm{d}t
\label{eq:a_int}
\end{align}
The integrands in Equations \ref{eq:loss_int} and \ref{eq:a_int} can both be computed using automatic differentiation. In MoSim, we treat $s(t)$ as the latent variable, and the dynamics network $g_\theta(s(t),t) = f_\theta(s(t),a(t)) + \sum_i \epsilon_\theta^i(s(t),a(t))$ is precisely composed of the predictor and corrector we mentioned earlier.

\subsection{Benchmark}
\label{subsec:benchmark}
While previous world models have typically been evaluated in reinforcement learning tasks, 
we propose a benchmark specifically designed to evaluate the predictive capability of the world model itself: continuous multi-step prediction of robotic motion. We use DeepMind Control \cite{tassa2018deepmindcontrolsuite} as the virtual environment for robotics, which is based on the physical dynamics simulation provided by MuJoCo \cite{todorov2012mujoco}.

We construct the test set in two ways: by generating random action sequences using Poisson sampling and by generating action sequences based on policies from pre-trained reinforcement learning agents. The former provides untrained, dispersed data across the action and state space, while the latter offers task-specific, meaningful data. We generate sufficiently long data sequences to ensure that both the random and policy-based data reach a steady state, and then randomly sample fragments of the desired horizon length for evaluation.

Our benchmark includes prediction horizons critical for existing models: 3 steps (for TD-MPC2\cite{hansen2024tdmpc2scalablerobustworld}), 16 steps (for DreamerV3\cite{hafner2023mastering}), and 100 steps (for long-term prediction). The world model is required to predict continuously for the specified horizon, given the initial state and the action sequence. For models that require a warm-up period to initialize latent variables and achieve better performance, our benchmark provides a certain number of ground truth steps as a condition before evaluation.

Finally, we compute the mean squared error loss between the model's predictions and the ground truth over the entire segment as a measure of prediction accuracy.

\subsection{Multi-Stage Training.}
\label{subsec:strategy}
\noindent {\bf }To fully utilize the model architecture proposed in Section \ref{subsec:arch}, we introduce a training strategy called multi-stage training. We first train the predictor without incorporating the corrector, assuming that it should capture the smooth, non-abrupt components of robotic dynamics, such as inertia, effects of gravity, and elastic forces. Once the predictor converges, we freeze it and train the corrector on top of it. At this stage, the corrector focuses on modeling the abrupt and unmodeled dynamics, attempting to bridge the gap between the actual dynamics and the smooth dynamics captured by the predictor. For more complex robots, we employ multiple correctors to perform stepwise refinement, with each one addressing the non-smooth dynamics that the previous level struggles to handle.

Compared to training the entire large network jointly from the start, multistage training process allows the model to first capitalize on the advantages of inductive biases to quickly learn the simpler and more fundamental parts of the task. Subsequently, the corrector, which lacks strong inductive biases, handles the more complex and task-specific components. Ablation studies in Section \ref{sec:results} demonstrate that models trained with the multistage training strategy achieve faster training speeds and improved final performance.


\section{Results}
\label{sec:results}
We evaluated MoSim in five robotic environments from the DM Control suite\cite{tunyasuvunakool2020}, as well as two additional environments: Panda\cite{menagerie2022github}, a robotic arm operating in a three-dimensional space, and Go2\cite{unitree_mujoco2021github, menagerie2022github}, a quadruped robot. We designed experiments to compare its predictive performance against current state-of-the-art world models, both in state space and latent space. The results demonstrate that MoSim achieved significant improvements in predictive capacity compared to previous world models. Additionally, we integrated MoSim into model-free algorithms, transforming them into model-based approaches, demonstrating that MoSim can be easily adapted to various model-free algorithms. Furthermore, we implemented zero-shot learning in specific tasks using MoSim-based reinforcement learning (RL) algorithms, enabling reinforcement learning without any real-world data, and explored the potential of few-shot learning in more complex tasks.

To provide a clear understanding of the model naming conventions used in our experiments, the suffix ``-r" indicates models trained on random datasets, where at each step, the action is independently sampled uniformly from the action space, ``-e" denotes models trained on experience datasets from Dreamer’s replay buffer, and ``-rm" refers to MoSim trained using a multistage approach on random data.

\subsection{Raw State Space Evaluation}
\label{subsec:state space comparsion}
We first compare the predictive capabilities of DreamerV3, whose backbone network is the Recurrent State-Space Model (RSSM)\cite{hafner2019rssm}, a variant of recurrent neural networks, with those of MoSim.

\noindent \textbf{RSSM Baseline.} We trained RSSM in two ways: one following the original DreamerV3 training method, and the other for an ablation experiment, following the same training approach as MoSim. We provided RSSM with 1-step and 5-step ground truth as initial conditions, based on our observation that the RSSM requires several initialization steps to prepare its latent variables, for better predictions.

\noindent \textbf{Generalization Ability.} As mentioned in the Introduction, we found that models trained on data collected in a random manner exhibit higher prediction accuracy and better generalization ability. We trained and compared both models(-r and -e) on datasets collected using both methods, and the results in Table \ref{tab:vsdreamer2} of the ablation experiments highlighted the superiority of the random data collection approach.

\noindent \textbf{Training Method Comparison.} We compared the impact of two different training methods on the results, demonstrating the effectiveness of the multistage training approach in terms of both time efficiency and overall performance.

\begin{table*}[t]
    \centering
    \begin{tabular}{@{}llccccccc}
        \toprule
        \multirow{2}{*}{Environment}& \multirow{2}{*}{Horizon}& \multicolumn{2}{c}{DreamerV3-e} &   \multicolumn{2}{c}{DreamerV3-r}& \multirow{2}{*}{MoSim-e} &\multirow{2}{*}{MoSim-r} &\multirow{2}{*}{MoSim-rm}\\
 & & 1-step& 5-step & 1-step&5-step&  & &\\
        \midrule
        Cheetah & 16 & 8.2538& 6.5270 & 0.8747&0.1925& 0.5342 &\textbf{0.1206} &/\\
        & 100 & 6.5507& 5.8468 & 0.4048&0.2297& 1.1876 &\textbf{0.2185} &/\\
        \midrule
        Reacher & 16 & 0.7370& 0.1081 & 0.7451&0.0972&  0.0037&\textbf{0.0005} &/\\
        & 100 & 0.7405& 0.1743 & 0.7532&0.0988&  0.0053&\textbf{0.0009} &/\\
        \midrule
        Acrobot  & 16 & 3.6390&  0.9356& 3.7423&0.1015&  /&\textbf{0.0001} &/\\
        & 100 & 9.7392&  4.8957& 9.7547&4.9276&  /&\textbf{0.1043} &/\\
        \midrule
        Panda& 16& /& /& 0.2396& 0.0434& /&\textbf{0.0010} &/\\
        & 100& /& /& 0.3367& 0.0971& /&\textbf{0.0043} &/\\
        \midrule
         Hopper& 16& /& /& 0.6406& 0.1114& /& 0.0743&\textbf{0.0375}\\
         & 100& /& /& 0.9239& 0.3199& /& 0.4049&\textbf{0.2507}\\
         \midrule
         Go2& 16& /& /& 1.0097& 0.3685& /& 0.0430&\textbf{0.0410}\\
         & 100& /& /& 0.9243& 0.4165& /& \textbf{0.1282}&0.1401\\
         \midrule
         Humanoid& 5& /& /& 5.2171& 1.3947& /& 0.9382&\textbf{0.6535}\\
         & 10& /& /& 6.0500& 1.8263& /& 1.5344&\textbf{1.0063}\\
         & 16& /& /& 6.5078& 2.1291& /& 1.9735&\textbf{1.2737}\\
         \bottomrule
    \end{tabular}
    \caption{MoSim versus DreamerV3. Evaluated on random policy datasets. Easy tasks. The suffix `-e' denotes models trained on experience data from Dreamer’s replay buffer, `-r' indicates models trained on random policy data, and `-rm' refers to MoSim trained on random policy data with the multistage approach.}
    \label{tab:vsdreamer1}
\end{table*}

In Table~\ref{tab:vsdreamer1}, We categorize the tasks into easy and hard, where easy tasks generally have smoother dynamics, making them more suited for predictor-based solutions. In these cases, we did not use the multistage training approach. On the other hand, hard tasks exhibit non-smooth dynamics, where multistage training significantly outperforms standard end-to-end training in terms of final performance. As illustrated in Figure~\ref{fig:ablation1}, the multistage approach also improves training efficiency and stability. Overall results indicate that MoSim consistently outperforms DreamerV3 across both easy and hard tasks, demonstrating its superior capability in handling diverse dynamics.

\begin{table*}[t]
    \centering
    \begin{tabular}{@{}llcccccc}
        \toprule
        \multirow{2}{*}{Environment}& \multirow{2}{*}{Horizon}& \multicolumn{2}{c}{DreamerV3-e}&  \multicolumn{2}{c}{DreamerV3-r}  & \multirow{2}{*}{MoSim-e}& \multirow{2}{*}{MoSim-r} \\
 & & 1-step& 5-step& 1-step& 5-step& & \\
        \midrule
        \multirow{2}{*}{Cheetah} & 16 & 13.1264& 4.8908& 18.4404 & 11.1325& \textbf{3.8841}& 5.6052\\
        & 100 & 13.8990& 9.4270& 18.6197 & 4.5623 & 13.0104& \textbf{3.8434}\\
        \midrule
        \multirow{2}{*}{Reacher} & 16 & 0.3430& 0.0514& 0.4670 & 0.04875 & 0.0027& \textbf{0.0005}\\
        & 100 & 0.3562& 0.0803& 0.4364 & 0.05712 & 0.0030& \textbf{0.0008}\\
        \midrule
        \multirow{2}{*}{Acrobot}  & 16 & 5.0681& 1.3900& 5.0681& 1.3900& /& \textbf{0.0121}\\
        & 100 & 16.8498& 11.1576& 16.8498& 11.1576& /& \textbf{1.2864}\\
        \bottomrule
    \end{tabular}
    \caption{MoSim versus DreamerV3. Evaluated on the TD-MPC2 policy dataset, which means the evaluation data is out-of-distribution (OOD) for all models. The results show that models trained on random datasets (-r) generally achieve lower prediction errors than model trained on experience datasets (-e), indicating a stronger generalization capability.}

    \label{tab:vsdreamer2}
\end{table*}

\subsection{Latent Space Evaluation}
\label{subsec:latent}
\noindent MoSim also exhibits strong prediction accuracy in the latent space of TD-MPC2. We utilized the signal task pre-trained TD-MPC2 model and leveraged its encoder to encode our predicted states into their latent space. We trained MoSim using tdmpc policy data and compared it against the model that generated the training and test data (TD-MPC2) in table \ref{tab:vstdmpc1}. In all environments, we performed three-step predictions, following the default prediction steps of TD-MPC2. We also do the comparison for random data in table \ref{tab:vstdmpc2}.

\setlength{\textfloatsep}{5pt}
\begin{table}
    \centering
    \begin{tabular}{@{}lcc@{}}
        \toprule
        Environment & TD-MPC2-e  & MoSim-e\\
        \midrule
        Humanoid &  \ 0.00011&\textbf{0.00009}\\
        Cheetah&  \ 0.0009&\textbf{0.0007}\\
        Reacher &  4.8101e-5&\textbf{2.9256e-7}\\

        \bottomrule
    \end{tabular}
    \caption{MoSim versus TD-MPC2. Evaluated on TD-MPC policy datasets. }
    \label{tab:vstdmpc1}
\end{table}

\begin{table}
    \centering
    \begin{tabular}{@{}lcc@{}}
        \toprule
        Environment & TD-MPC2-e  & MoSim-rm\\
        \midrule
        Humanoid &  0.0050&\textbf{0.0020}\\
        Cheetah&  0.0011&\textbf{0.0003}\\
        Reacher& 9.9229e-5&\textbf{7.0000e-7}\\
        \bottomrule
    \end{tabular}
    \caption{MoSim versus TD-MPC2. Evaluated on random datasets.}
    \label{tab:vstdmpc2}
\end{table}

\vspace{-0.2em}

\subsection{Stochasticity}
\label{Stochasticity}
While MoSim assumes deterministic dynamics, we examine its robustness to observation noise and its capability to reliably sample multiple future trajectories.

Table \ref{tab:Noise training} shows that MoSim remains robust to noise in training data while maintaining consistent or even achieving better performance. 

\begin{table}[htbp]
\centering
\resizebox{\linewidth}{!}{%
\begin{tabular}{@{}lccc@{}}
\toprule
Prediction MSE Loss& DreamerV3 & MoSim & \makecell{MoSim\\w/ noise $\mathcal{N}\sim(0,0.01^{2})$} \\
\midrule
Reacher(100 steps) & 0.0988 & 0.0009 & 0.0008 \\
\midrule
Panda(100 steps) & 0.0971 & 0.0043 & 0.0042 \\
\midrule
Go2(100 steps)& 0.4165 & 0.1282 & 0.2111 \\
\bottomrule

\end{tabular}
}
\caption{Prediction accuracy of MoSim trained on noisy data, compared to its of DreamerV3 and normal MoSim.}
\label{tab:Noise training}
\end{table}

\begin{table}
\centering
\resizebox{\linewidth}{!}{%
\begin{tabular}{@{}l|ll@{}}
\toprule
\multirow{2}{*}{\makecell{Lyapunov Characteristic Exponent (LCE)\\ $\delta=10^{-5},T=1000,\text{\ avg on 2000 trajectories}$}}
 & MuJoCo & MoSim \\
 & 1.1738&  1.1728\\
\bottomrule
\end{tabular}
}
\caption{Numerical results of LCE of MuJoCo and MoSim}
\label{tab:Noise Sampling}
\end{table}

 For a self-consistent chaotic system such as Acrobot, multiple future trajectories sampling can be reached through ensemble sampling. Table \ref{tab:Noise Sampling} shows that MoSim's Lyapunov characteristic exponent aligns with MuJoCo, confirming the reliability of ensemble sampling.

\subsection{Zero- and Few-Shot Reinforcement Learning}
\label{subsec:zero_shot}
Incorporating MoSim enables us to seamlessly transform any model-free reinforcement learning algorithm into a model-based approach, leveraging the predictive capabilities of MoSim's dynamics model. In this section, we integrate MoSim with TQC \cite{kuznetsov2020controllingoverestimationbiastruncated} and SAC \cite{haarnoja2018softactorcriticoffpolicymaximum}, introducing three distinct integration strategies, each offering varying levels of data efficiency.

\noindent \textbf{Zero-Shot Learning.} By capitalizing on the robust generalization capabilities and predictive performance of MoSim, we employed a MoSim model trained on randomized data as a surrogate for the {dm\_control} environment. This enabled the agent to interact directly with the world model for training, achieving scores on three {dm\_control} benchmark tasks that closely approximate those of the real environment.

\begin{figure*}[htbp]
    \centering

    \begin{subfigure}{0.32\linewidth}
        \centering
        \includegraphics[width=\linewidth]{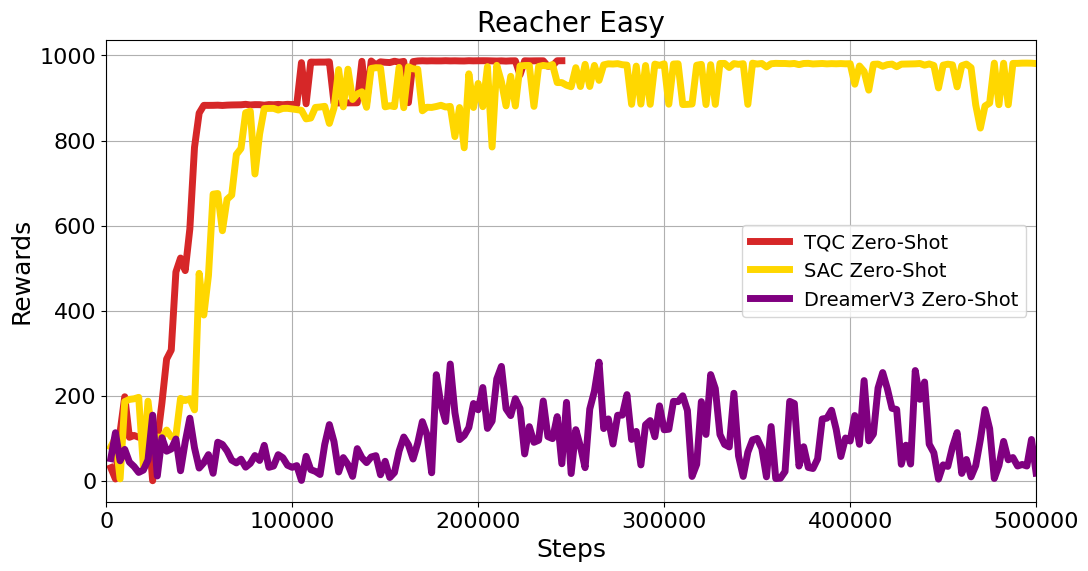}
        \caption{Zero-Shot Reacher-Easy}
        \label{fig:result1}

    \end{subfigure}
    \hfill
    \begin{subfigure}{0.32\linewidth}
        \centering
        \includegraphics[width=\linewidth]{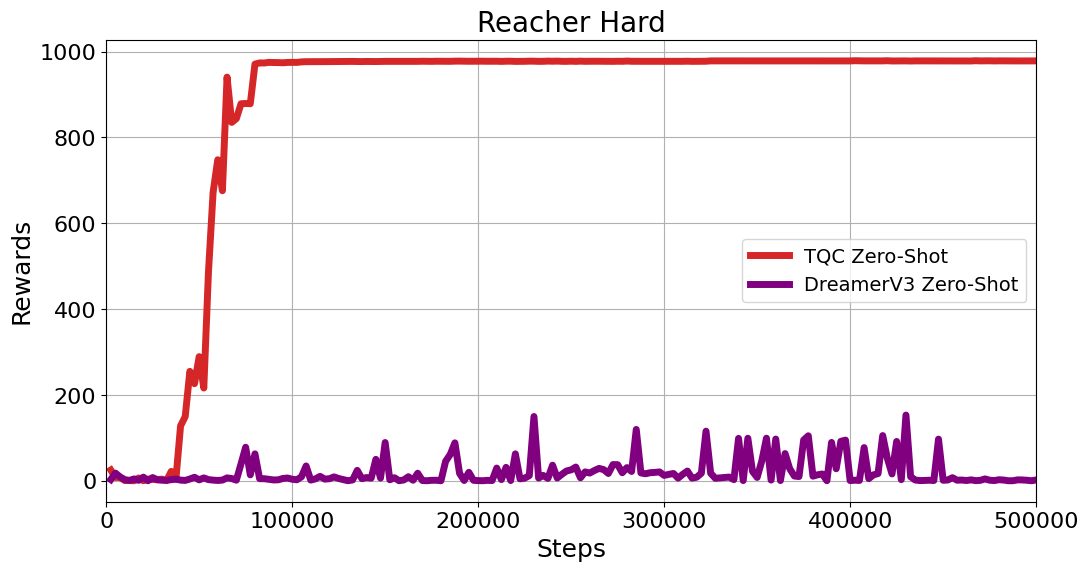}
        \caption{Zero-Shot Reacher-Hard}
        \label{fig:result2}

    \end{subfigure}
    \hfill
    \begin{subfigure}{0.32\linewidth}
        \centering
        \includegraphics[width=\linewidth]{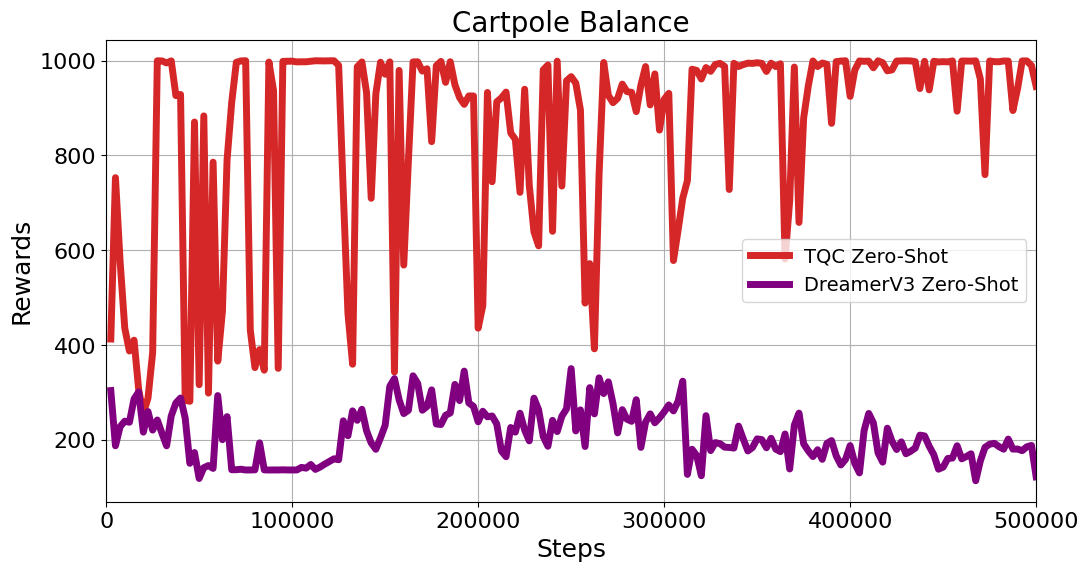}
        \caption{Zero-Shot Cartpole-Balance}
        \label{fig:result3}

    \end{subfigure}
     \begin{subfigure}{0.32\linewidth}
        \centering
        \includegraphics[width=\linewidth]{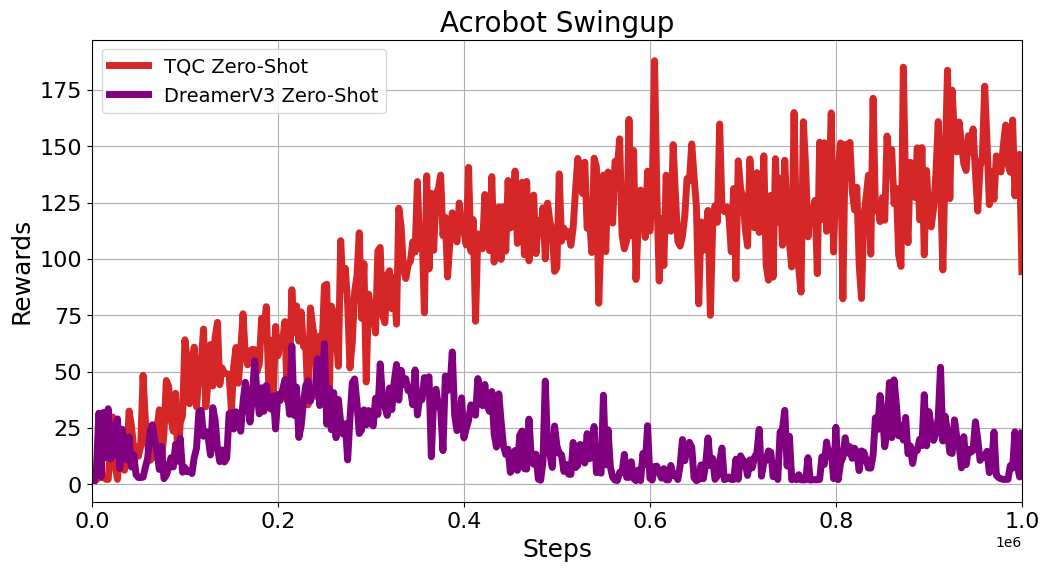}
        \caption{Zero-Shot Acrobot-SwingUp}
        \label{fig:result4}

    \end{subfigure}
    \hspace{0.01\linewidth}
    \begin{subfigure}{0.32\linewidth}
        \centering
        \includegraphics[width=\linewidth]{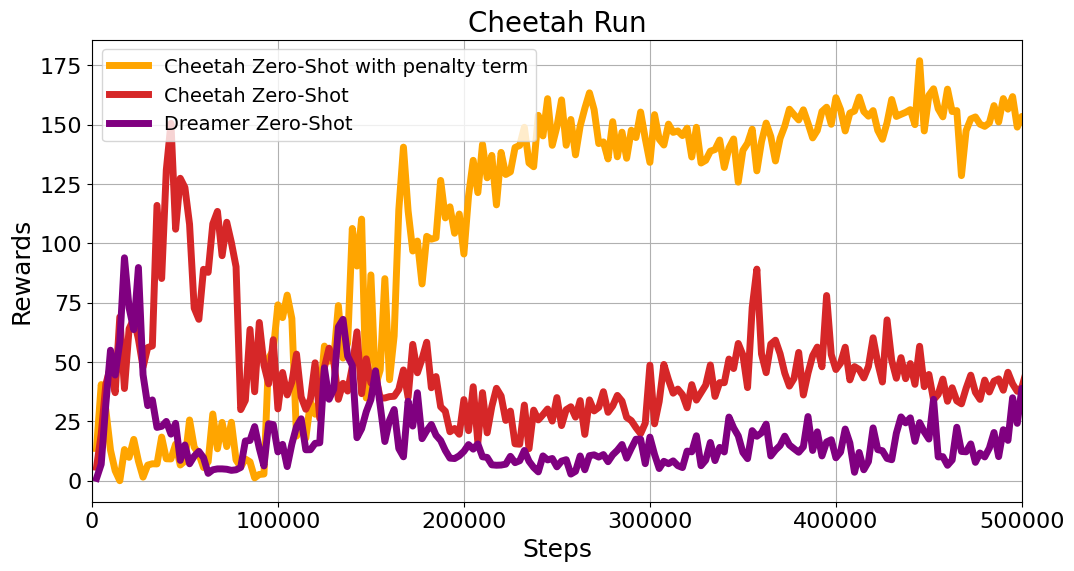}
        \caption{Zero-Shot Cheetah-Run}
        \label{fig:result5}
    \end{subfigure}
    \hspace{0.01\linewidth}
    \begin{subfigure}{0.32\linewidth}
        \centering
        \includegraphics[width=\linewidth]{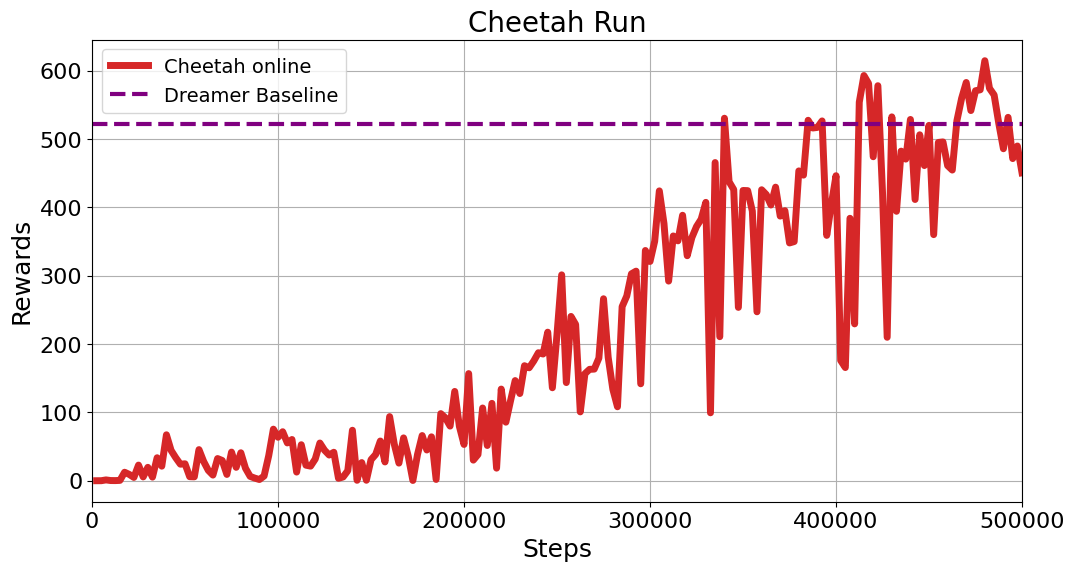}
        \caption{online Cheetah-Run}
        \label{fig:result6}
    \end{subfigure}

\end{figure*}

When trying the Cheetah-Run task, two significant challenges have hindered progress. First, the absolute upper limit of the model in predictive capability is insufficient. In the Cheetah-Run task, the MoSim model is able to maintain stable predictions only up to approximately 100 steps. However, it was observed that the existing model-free algorithms fail to successfully learn the Cheetah-Run task within the original environment when the time limit is set to 100. Additionally, as training progresses, a severe distribution shift in the training data becomes apparent, eventually exceeding MoSim's generalization capacity. At this stage, due to increasing prediction inaccuracies, the RL training plateaus (around a score of 100 for the Cheetah-Run task) and may even begin to degrade. These two issues have impeded further progress in MoSim's zero-shot learning for the Cheetah-Run task.

\noindent \textbf{Few-Shot Learning.} To address the two challenges encountered in zero-shot learning, we propose a few-shot learning strategy. Instead of solely relying on MoSim to replace the virtual environment, we periodically collect real environment data during training. Specifically, for every 5,000 virtual environment steps, we gather 1,000 steps of real environment data. This data is then used to train MoSim by sampling the real dataset every 100 virtual steps, helping to mitigate the distribution shift that occurs during training. Additionally, to address the issue of limited prediction horizon, we initialize the virtual environment with states sampled from the real environment replay buffer at each reset, rather than starting from the initial state every time. This aims to mimic the effect of having a time limit of 1,000 steps in the real environment. The results indicate that, unlike zero-shot learning, where performance plateaus and then deteriorates, reinforcement learning stabilizes and fluctuates around a score of 100. However, the curve still fails to improve and remains at a relatively low score.

\subsection{How Much Better Do We Need It To Be?}
We investigated the minimum environment time limit required for the original TQC algorithm to achieve its baseline performance under precise predictions. We varied the time limit in the dm\_control environment and evaluated its impact on the algorithm's performance. Table \ref{tab:zero-shot requirements} implies that for the TQC algorithm to achieve offline RL without performance degradation, the World Model must possess the capability to continuously predict over the minimum step limit length across any data distribution. This remains a critical goal we aim to achieve in the long term.

It is worth noting that the ``zero-shot horizon requirement" in Table \ref{tab:zero-shot requirements} was actually an upper bound, as it was calculated with only one model-free RL algorithm. There exist algorithms that can make this bound significantly tighter. For example, MPC requires only a 50-step horizon for the Acrobot-SwingUp task\cite{AcrobotMPC}. We provide a more detailed discussion of this in the Appendix.

\begin{table}
    \centering
    \resizebox{\linewidth}{!}{%
        \begin{tabular}{ccccc}
            \toprule
            Task Name & Cheetah & Reacher-Easy & Reacher-Hard & Acrobot \\
            \midrule
            Horizon & 500 & 30 & 30 & 800 \\
            Achieved & 60 & 30 & 30 & 60 \\
            \bottomrule
        \end{tabular}
    }
    \caption{Zero-Shot Horizon Requirements. The ``Horizon" indicates the minimum number of prediction steps required for complete zero-shot performance, while ``Achieved" indicates the current number of steps MoSim has successfully reached.}
    \label{tab:zero-shot requirements}
\end{table}

\subsection{On-Policy World Model for RL}

In previous successful world models, the model was updated in an on-policy and online fashion, where the world model was trained using samples from a collected replay buffer at every real environment step. The trained data was then immediately used as the starting point for the model's predictions, creating a continuous loop. However, this approach inherently involves frequent interactions with the real environment, making it challenging to reduce the number of real-world interactions.  MoSim adopted a similar training strategy: every 100 real environment steps, a batch of data was sampled from the replay buffer to train the model. For the subsequent 100 steps, the model used sampled states from the replay buffer as initial states to predict one step ahead and train the policy. With this training strategy, MoSim outperformed DreamerV3\footnote{All DreamerV3 reinforcement learning results reported in this paper are obtained directly from the open-source implementation of the DreamerV3 baseline provided in the TD-MPC2 repository: \url{https://github.com/nicklashansen/tdmpc2}} within the 500k time step limit on the cheetah task~\ref{fig:result3}.To provide a contrast, we used a pre-trained MoSim model as the world model, which was pre-trained on random data. In this case, the model predicted one step ahead in each step but was never updated with new data. As shown in Figure~\ref{fig:offline}, while the model initially showed improvement, the prediction quality eventually declined because the world model failed to generalize to the current data distribution, ultimately leading to a collapse in training performance.

Although MoSim demonstrated strong performance on the cheetah task with an on-policy training strategy, we believe that few-shot learning remains a promising and worthwhile direction to pursue. Achieving purely zero-shot performance on locomotion tasks is nearly impossible, as the data distribution can vary significantly across different policies and training stages. Additionally, the model's predictive capabilities are inherently limited in such tasks. Thus, leveraging real data to assist training can be essential. Few-shot learning provides a feasible and meaningful solution, as it significantly reduces the need for interactions with the real environment, thereby improving data efficiency and making the approach more practical for complex tasks.

\subsection{Let Model Know When It Doesn't Know}
To address the issue mentioned in Section \ref{subsec:zero_shot} where performance degrades during later stages of training due to distribution shift, our aim is to enable the model to recognize when it is uncertain, allowing exploration to avoid areas where the world model tends to generalize poorly and produce inaccurate predictions. Specifically, we use a residual flow with a Gaussian base distribution to fit the distribution of the MoSim training set. We then incorporate the probability density of each data point as a penalty term in the reinforcement learning reward. In this way, when a data point has a low probability density under the residual flow (indicating that the world model is unfamiliar with that region of data, i.e., it has encountered a distribution shift), then the policy is discouraged from taking that action. As shown in the figure\ref{fig:result5} , after adding the penalty term to the reward, the zero-shot Cheetah training reward (without the penalty term) no longer declines after a period of training and even achieves a higher score compared to the original maximum. This result provides an initial indication and potential future direction for addressing the distribution shift.(A detailed explanation of this method is provided in the Appendix~\ref{appendix:residual_flow}.)

\subsection{How Predictive Capacity Affects RL}
\begin{figure}
    \centering
    \includegraphics[width=1\linewidth,width=0.75\linewidth]{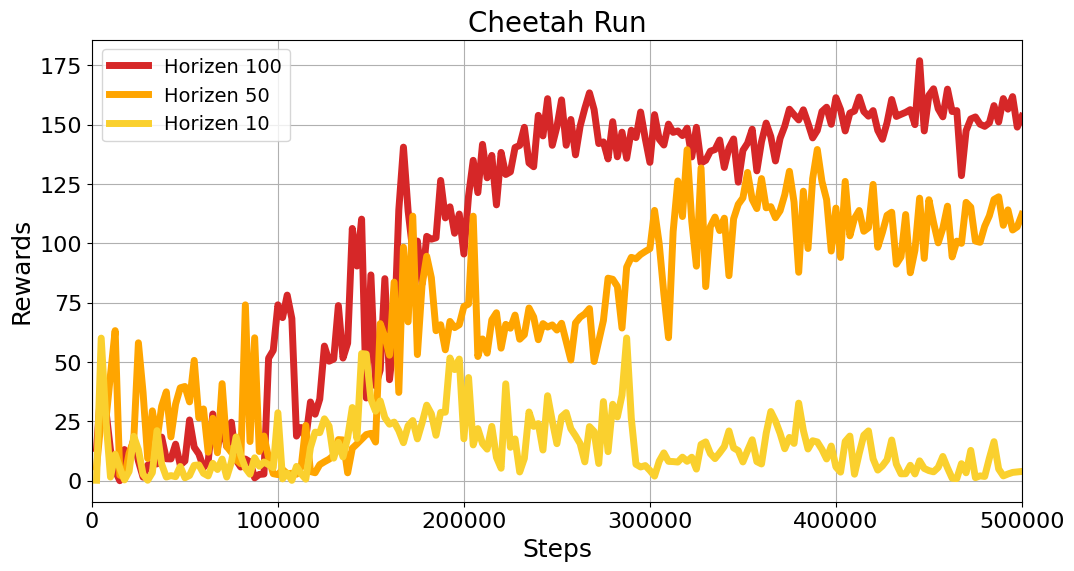}
  \caption{Policy learning with different prediction horizen}
    \label{fig:horizen_c}
\end{figure}

To explore the impact of the predictive capacity of the world model on reinforcement learning, we compare the final performance of policies trained with different prediction horizons: 10, 50, and 100. As shown in Figure \ref{fig:horizen_c}, the results indicate that as the prediction horizon decreases, the final performance of the policy also decreases. This highlights the importance of the prediction capacity of the world model for reinforcement learning.

\subsection{Ablation Study}
We conducted two ablation studies to evaluate the effectiveness of the network architecture and training strategies discussed in Section \ref{sec:method}.

\begin{figure}
    \centering
    \includegraphics[width=1\linewidth,width=\linewidth]{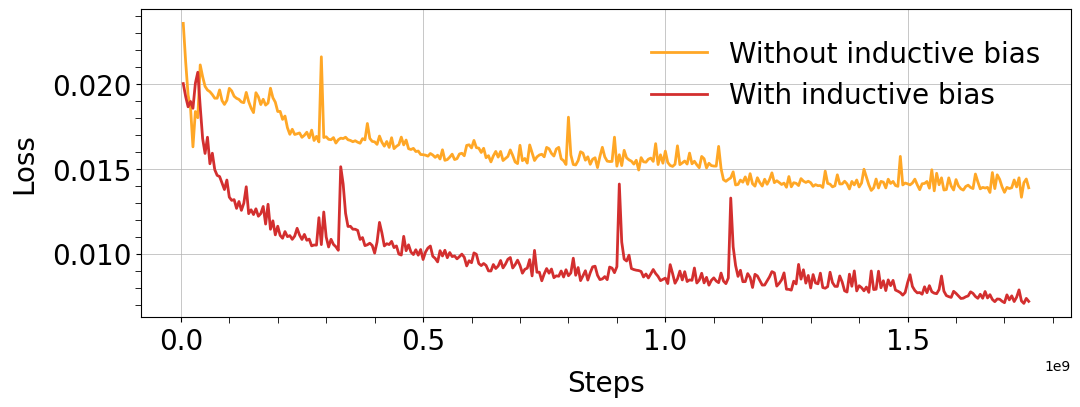}
  \caption{Ablation study of inductive bias on Hopper-Hop.}
    \label{fig:ablation1}
\end{figure}

\noindent \textbf{Inductive Bias of Rigid Body Dynamics:} We compared the predictor model with a standard ResNet of nearly identical parameter count to demonstrate the necessity of introducing inductive bias. The results, shown in Figure~\ref{fig:ablation1}, illustrate the clear benefits of inductive bias in terms of training speed and overall performance.

\begin{figure}
    \centering
    \includegraphics[width=\linewidth,width=
    \linewidth]{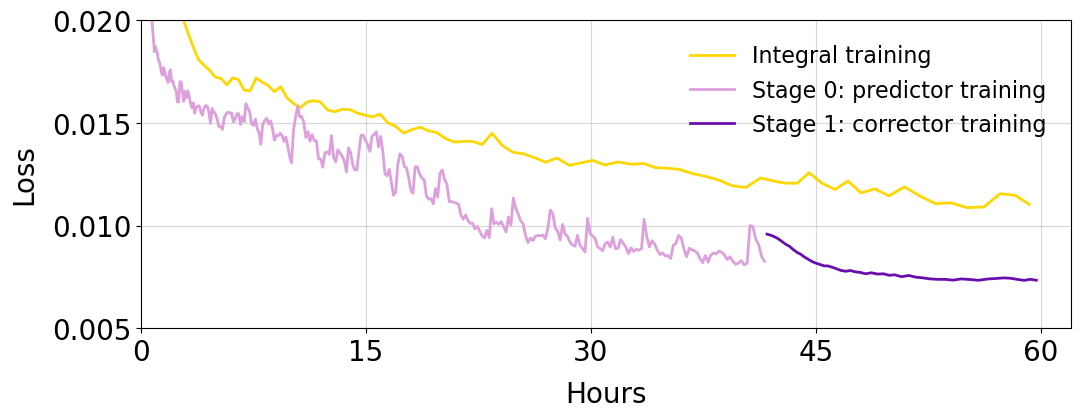}
  \caption{Ablation study of training method on Hopper-Hop.}
    \label{fig:ablation2}
\end{figure}

\noindent \textbf{Multistage Training:} We conducted an ablation study using a predictor model and a corrector model with identical parameter counts. Both models were evaluated under full training and multi-stage training regimes. The results, presented in Figure~\ref{fig:ablation2}, demonstrate the effectiveness of the multi-stage training approach.

\setlength{\textfloatsep}{5pt}

\section{Additional Related Works}
\label{sec:additional_related_works}

\noindent \textbf{Differentiable Simulation.} 
Recent differentiable physics engines~\cite{heiden2021neuralsim, hu2019chainqueen} offer end-to-end gradients via hand-coded rigid/soft-body models. In contrast, our data-driven approach minimizes manual force/contact design, using a structural predictor with trainable correctors.

\medskip
\noindent \textbf{Neural ODE Usage.}
Neural ODEs have been applied to time-series modeling~\cite{rubanova2019latentode}, fluid simulation~\cite{massaroli2020dissecting}, system ID~\cite{khodayi-mehr2020deepmechanicalode}, and generative modeling~\cite{yildiz2019ode2vae}, and have inspired physics-based networks~\cite{finzi2020simplifying}. Building on this, we use a neural ODE backbone with rigid-body insight and residual correctors for robust long-horizon prediction in robotic control.

\medskip
\noindent \textbf{Parameter Identification.} 
Classical optimal control refines simulations via identifying parameters like mass or friction~\cite{morris2011parameter}. In contrast, our method learns a scalable dynamics model from data, capturing smooth and non-smooth effects without manual engineering.

\medskip
\noindent {\bf World Models and Model-Based RL.} One major application of Existing world models are mainly used in reinforcement learning, where they predict future states in explicit or latent space as action-conditioned models~\cite{deisenroth2011pilco, chua2018pets, hafner2020mastering, hafner2023mastering, hafner2019dream, janner2019mbpo, theodorou2010mpc, schrittwieser2020muzero, hansen2024tdmpc2scalablerobustworld}. Such inner-model-based methods are called model-based RL and offer key advantages over model-free approaches by leveraging model predictions.

\begin{enumerate}
    \item \textbf{Pixel-Based Video Generation World Models.} 
    Another line of work predicts future frames from actions for pixel-based policy learning~\cite{ha2018worldmodels, hafner2019planet}. Models like DreamerV2/V3~\cite{hafner2023mastering} learn latent visual representations but struggle with fine-grained physical interactions like contact.
    \item \textbf{Data Efficient Learning}: RL algorithms can learn from the data generated by a trained model's predictions (or ``dreams"), effectively treating the model as a form of data augmenter.\cite{deisenroth2011pilco, azizzadenesheli2018atari, xu2023feasibilitycrosstasktransfermodelbased, hua2023simpleemergentactionrepresentations}
    \item \textbf{Planning Capability}: Many real-world tasks demand long-term planning, where models predict outcomes over action sequences and search for optimal ones~\cite{theodorou2010mpc, schrittwieser2020muzero, silver2018alphazero}. Planning has shown great success, e.g., AlphaGo~\cite{silver2016alphago}, but real-world dynamics are far harder to predict than a Go board. Enabling real-world planning first requires a predictive world model.
\end{enumerate}

\section{Conclusion}
\label{sec:conclusion}

In this work, we focus on evaluating and improving a world model's direct predictive capability. The resulted model, MoSim, significantly outperforms the world models in direct state prediction. We show, for the first time, that when a world model's prediction horizon and accuracy are sufficient, it is possible to train or search for a new policy purely in the predicted space in a zero-shot manner. However, a notable gap remains in achieving few-shot learning or even zero-shot learning in many cases. Our findings highlight that world models for motion dynamics is a promising direction for developing more versatile and capable embodied systems.

{
    \small
    \bibliographystyle{unsrt}
    \bibliography{main}

\begin{thebibliography}{10}

\bibitem{lecun2022path}
Yann LeCun.
\newblock A path towards autonomous machine intelligence version 0.9. 2, 2022-06-27.
\newblock {\em Open Review}, 62(1):1--62, 2022.

\bibitem{sutton1991dyna}
Richard~S Sutton.
\newblock Dyna, an integrated architecture for learning, planning, and reacting.
\newblock {\em ACM Sigart Bulletin}, 2(4):160--163, 1991.

\bibitem{henaff2017model}
Mikael Henaff, William~F Whitney, and Yann LeCun.
\newblock Model-based planning with discrete and continuous actions.
\newblock {\em arXiv preprint arXiv:1705.07177}, 2017.

\bibitem{ha2018world}
David Ha and J{\"u}rgen Schmidhuber.
\newblock World models.
\newblock {\em arXiv preprint arXiv:1803.10122}, 2018.

\bibitem{hansen2022temporaldifferencelearningmodel}
Nicklas Hansen, Xiaolong Wang, and Hao Su.
\newblock Temporal difference learning for model predictive control, 2022.

\bibitem{hansen2024tdmpc2scalablerobustworld}
Nicklas Hansen, Hao Su, and Xiaolong Wang.
\newblock Td-mpc2: Scalable, robust world models for continuous control, 2024.

\bibitem{hafner2019dream}
Danijar Hafner, Timothy Lillicrap, Jimmy Ba, and Mohammad Norouzi.
\newblock Dream to control: Learning behaviors by latent imagination.
\newblock {\em arXiv preprint arXiv:1912.01603}, 2019.

\bibitem{hafner2020mastering}
Danijar Hafner, Timothy Lillicrap, Mohammad Norouzi, and Jimmy Ba.
\newblock Mastering atari with discrete world models.
\newblock {\em arXiv preprint arXiv:2010.02193}, 2020.

\bibitem{hafner2023mastering}
Danijar Hafner, Jurgis Pasukonis, Jimmy Ba, and Timothy Lillicrap.
\newblock Mastering diverse domains through world models.
\newblock {\em arXiv preprint arXiv:2301.04104}, 2023.

\bibitem{chen2018neural}
Ricky~TQ Chen, Yulia Rubanova, Jesse Bettencourt, and David~K Duvenaud.
\newblock Neural ordinary differential equations.
\newblock {\em Advances in neural information processing systems}, 31, 2018.

\bibitem{goldstein2001classical}
Herbert Goldstein, Charles Poole, and John Safko.
\newblock {\em Classical Mechanics}.
\newblock Pearson, 3rd edition, 2001.

\bibitem{todorov2012mujoco}
Emanuel Todorov, Tom Erez, and Yuval Tassa.
\newblock Mujoco: A physics engine for model-based control.
\newblock In {\em 2012 IEEE/RSJ International Conference on Intelligent Robots and Systems (IROS)}, pages 5026--5033, 2012.

\bibitem{he2016deep}
Kaiming He, Xiangyu Zhang, Shaoqing Ren, and Jian Sun.
\newblock Deep residual learning for image recognition.
\newblock In {\em Proceedings of the IEEE conference on computer vision and pattern recognition}, pages 770--778, 2016.

\bibitem{golub2013matrix}
Gene~H. Golub and Charles~F. Van~Loan.
\newblock {\em Matrix Computations}.
\newblock Johns Hopkins University Press, Baltimore, MD, 4th edition, 2013.
\newblock Cholesky decomposition is covered in Chapter 4.

\bibitem{euler1768integralis}
Leonhard Euler.
\newblock {\em Institutiones Calculi Integralis}, volume~1.
\newblock Bousquet, 1768.

\bibitem{butcher2003runge}
J.C. Butcher.
\newblock {\em Numerical Methods for Ordinary Differential Equations}.
\newblock John Wiley \& Sons, Chichester, UK, 2003.

\bibitem{kutta1901runge}
W.~Kutta.
\newblock Beitrag zur näherungsweisen integration totaler differentialgleichungen.
\newblock {\em Zeitschrift für Mathematik und Physik}, 46:435--453, 1901.

\bibitem{dormand1980runge}
J.R. Dormand and P.J. Prince.
\newblock A family of embedded runge-kutta formulae.
\newblock {\em Journal of Computational and Applied Mathematics}, 6:19--26, 1980.

\bibitem{tassa2018deepmindcontrolsuite}
Yuval Tassa, Yotam Doron, Alistair Muldal, Tom Erez, Yazhe Li, Diego de~Las~Casas, David Budden, Abbas Abdolmaleki, Josh Merel, Andrew Lefrancq, Timothy Lillicrap, and Martin Riedmiller.
\newblock Deepmind control suite, 2018.

\bibitem{tunyasuvunakool2020}
Saran Tunyasuvunakool, Alistair Muldal, Yotam Doron, Siqi Liu, Steven Bohez, Josh Merel, Tom Erez, Timothy Lillicrap, Nicolas Heess, and Yuval Tassa.
\newblock dm\_control: Software and tasks for continuous control.
\newblock {\em Software Impacts}, 6:100022, 2020.

\bibitem{menagerie2022github}
Kevin Zakka, Yuval Tassa, and {MuJoCo Menagerie Contributors}.
\newblock {MuJoCo Menagerie: A collection of high-quality simulation models for MuJoCo}, 2022.

\bibitem{unitree_mujoco2021github}
Unitree Robotics.
\newblock {Unitree Mujoco}, 2021.

\bibitem{hafner2019rssm}
Danijar Hafner, Timothy Lillicrap, Mohammad Norouzi, and Jimmy Ba.
\newblock Learning latent dynamics for planning from pixels.
\newblock {\em arXiv preprint arXiv:1912.01603}, 2019.

\bibitem{kuznetsov2020controllingoverestimationbiastruncated}
Arsenii Kuznetsov, Pavel Shvechikov, Alexander Grishin, and Dmitry Vetrov.
\newblock Controlling overestimation bias with truncated mixture of continuous distributional quantile critics, 2020.

\bibitem{haarnoja2018softactorcriticoffpolicymaximum}
Tuomas Haarnoja, Aurick Zhou, Pieter Abbeel, and Sergey Levine.
\newblock Soft actor-critic: Off-policy maximum entropy deep reinforcement learning with a stochastic actor, 2018.

\bibitem{AcrobotMPC}
Mingyuan Zhong, Mikala Johnson, Yuval Tassa, Tom Erez, and Emanuel Todorov.
\newblock Value function approximation and model predictive control.
\newblock In {\em 2013 IEEE Symposium on Adaptive Dynamic Programming and Reinforcement Learning (ADPRL)}, pages 100--107, 2013.

\bibitem{heiden2021neuralsim}
Enric Heiden, Daniel Millard, Erwin Coumans, Gaurav~S. Sukhatme, and Konstantinos Tsotsos.
\newblock Neuralsim: Augmenting differentiable simulators with neural networks.
\newblock {\em International Conference on Learning Representations (ICLR)}, 2021.

\bibitem{hu2019chainqueen}
Yifei Hu, Dylan Jacobs, Boyuan~Chen Huang, et~al.
\newblock Chainqueen: A real-time differentiable physical simulator for soft robotics.
\newblock {\em Robotics: Science and Systems (RSS)}, 2019.

\bibitem{rubanova2019latentode}
Yulia Rubanova, Ricky~TQ Chen, and David Duvenaud.
\newblock Latent ordinary differential equations for irregularly-sampled time series.
\newblock {\em Advances in Neural Information Processing Systems (NeurIPS)}, 32, 2019.

\bibitem{massaroli2020dissecting}
Stefano Massaroli, Michael Poli, Jinkyoo Park, Atsushi Yamashita, and Hajime Nakamura.
\newblock Dissecting neural odes.
\newblock {\em International Conference on Machine Learning (ICML)}, pages 1--11, 2020.

\bibitem{khodayi-mehr2020deepmechanicalode}
Roohollah Khodayi-Mehr, Andreas Walch, and P-S Koutsourelakis.
\newblock Deep learning of mechanical systems: A neural-ode approach.
\newblock {\em Computers \& Structures}, 235:106--150, 2020.

\bibitem{yildiz2019ode2vae}
Caglar~G. Yildiz, Markus Heinonen, and Harri L{\"a}hdesm{\"a}ki.
\newblock Ode2vae: Deep generative second order odes with bayesian neural networks.
\newblock {\em Advances in Neural Information Processing Systems (NeurIPS)}, 32, 2019.

\bibitem{finzi2020simplifying}
Marc Finzi, Ke~Alexander Wang, Andrew Xu, and Andrew~Gordon Wilson.
\newblock Simplifying hamiltonian and lagrangian neural networks via explicit constraints.
\newblock {\em Advances in Neural Information Processing Systems (NeurIPS)}, 33:13880--13890, 2020.

\bibitem{morris2011parameter}
Karline Morris and Anil~V Rao.
\newblock Parameter identification for optimal control problems with applications in aerospace engineering.
\newblock {\em Optimization and Engineering}, 12(1):123--144, 2011.

\bibitem{deisenroth2011pilco}
Marc~P. Deisenroth and Carl~E. Rasmussen.
\newblock Pilco: A model-based and data-efficient approach to policy search.
\newblock In {\em Proceedings of the International Conference on Machine Learning (ICML)}, pages 465--472, 2011.

\bibitem{chua2018pets}
Kurtland Chua, Roberto Calandra, Rowan McAllister, and Sergey Levine.
\newblock Deep reinforcement learning in a handful of trials using probabilistic dynamics models.
\newblock In {\em Advances in Neural Information Processing Systems (NeurIPS)}, pages 4754--4765, 2018.

\bibitem{janner2019mbpo}
Michael Janner, Justin Fu, Marvin Zhang, and Sergey Levine.
\newblock When to trust your model: Model-based policy optimization.
\newblock In {\em Advances in Neural Information Processing Systems (NeurIPS)}, pages 12519--12530, 2019.

\bibitem{theodorou2010mpc}
Evangelos Theodorou, Jonas Buchli, and Stefan Schaal.
\newblock Model predictive path integral control: From theory to parallel computation.
\newblock {\em The International Journal of Robotics Research}, 31(2):160--187, 2010.

\bibitem{schrittwieser2020muzero}
Julian Schrittwieser and et~al.
\newblock Mastering atari, chess, go, and shogi by planning with a learned model.
\newblock {\em Nature}, 588(7839):604--609, 2020.

\bibitem{ha2018worldmodels}
David Ha and J{\"u}rgen Schmidhuber.
\newblock World models.
\newblock {\em arXiv preprint arXiv:1803.10122}, 2018.

\bibitem{hafner2019planet}
Danijar Hafner, Timothy Lillicrap, Ian Fischer, Ruben Villegas, David Ha, Honglak Lee, and James Davidson.
\newblock Learning latent dynamics for planning from pixels.
\newblock In {\em International Conference on Machine Learning (ICML)}, pages 2555--2565, 2019.

\bibitem{azizzadenesheli2018atari}
Michael~J. Azizzadenesheli, Emma Brunskill, and Animashree Anandkumar.
\newblock Model-based reinforcement learning for atari.
\newblock In {\em International Conference on Learning Representations (ICLR)}, 2018.

\bibitem{xu2023feasibilitycrosstasktransfermodelbased}
Yifan Xu, Nicklas Hansen, Zirui Wang, Yung-Chieh Chan, Hao Su, and Zhuowen Tu.
\newblock On the feasibility of cross-task transfer with model-based reinforcement learning, 2023.

\bibitem{hua2023simpleemergentactionrepresentations}
Pu~Hua, Yubei Chen, and Huazhe Xu.
\newblock Simple emergent action representations from multi-task policy training, 2023.

\bibitem{silver2018alphazero}
David Silver, Thomas Hubert, Julian Schrittwieser, Ioannis Antonoglou, Matthew Lai, Arthur Guez, Marc Lanctot, Laurent Sifre, Dharshan Kumaran, Thore Graepel, Timothy Lillicrap, Karen Simonyan, and Demis Hassabis.
\newblock A general reinforcement learning algorithm that masters chess, shogi, and go through self-play.
\newblock {\em Science}, 362(6419):1140--1144, 2018.

\bibitem{silver2016alphago}
David Silver, Aja Huang, Chris~J. Maddison, Arthur Guez, Laurent Sifre, George Van Den~Driessche, Julian Schrittwieser, Ioannis Antonoglou, Veda Panneershelvam, Marc Lanctot, Sander Dieleman, Dominik Grewe, John Nham, Nal Kalchbrenner, Ilya Sutskever, Timothy Lillicrap, Madeleine Leach, Koray Kavukcuoglu, Thore Graepel, and Demis Hassabis.
\newblock Mastering the game of go with deep neural networks and tree search.
\newblock {\em Nature}, 529(7587):484--489, 2016.

\end{thebibliography}
}

\clearpage
\clearpage
\setcounter{page}{1}
\renewcommand{\thesection}{\Alph{section}}
\renewcommand{\thefigure}{\Roman{figure}}
\renewcommand{\thetable}{\Roman{table}}

\setcounter{section}{0}
\setcounter{figure}{0}
\setcounter{table}{0}

\section*{Appendix}

\section{Architecture}

\begin{figure}[htbp]
    \centering
    \begin{subfigure}[b]{0.2\textwidth}
        \includegraphics[width=\textwidth]{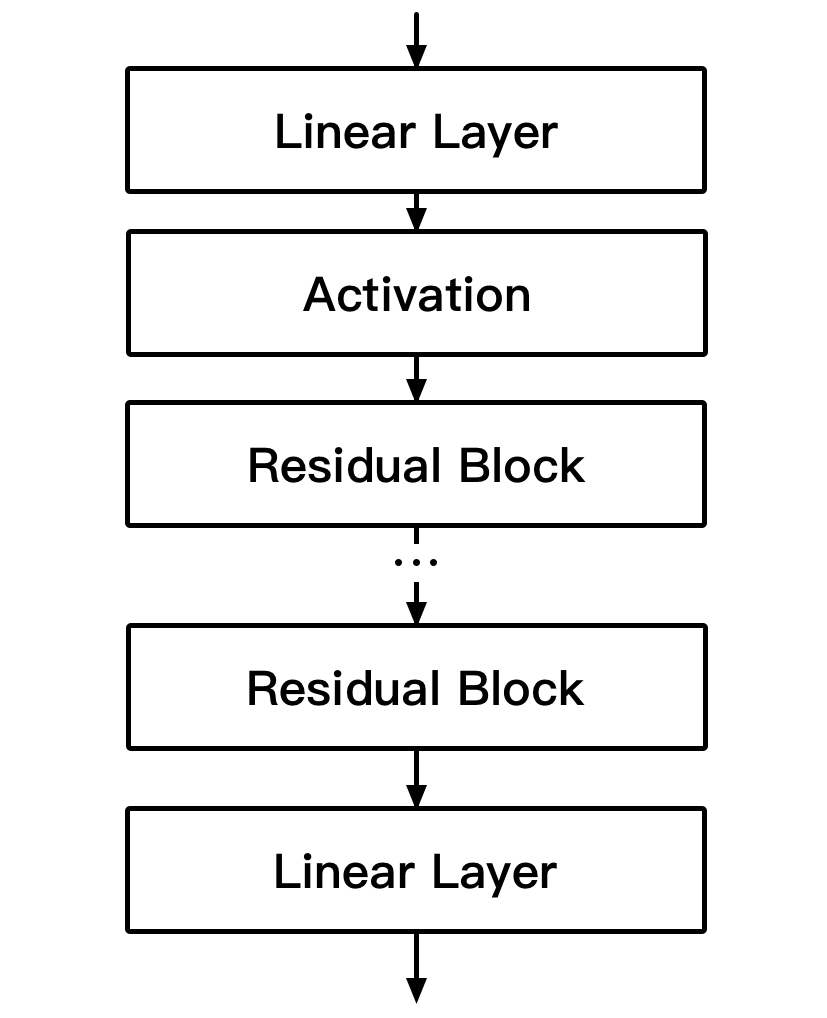}
        \caption{Residual Network}
        \label{fig:residual_network}
    \end{subfigure}
    \hspace{0.5em}
    \begin{subfigure}[b]{0.2\textwidth}
        \includegraphics[width=\textwidth]{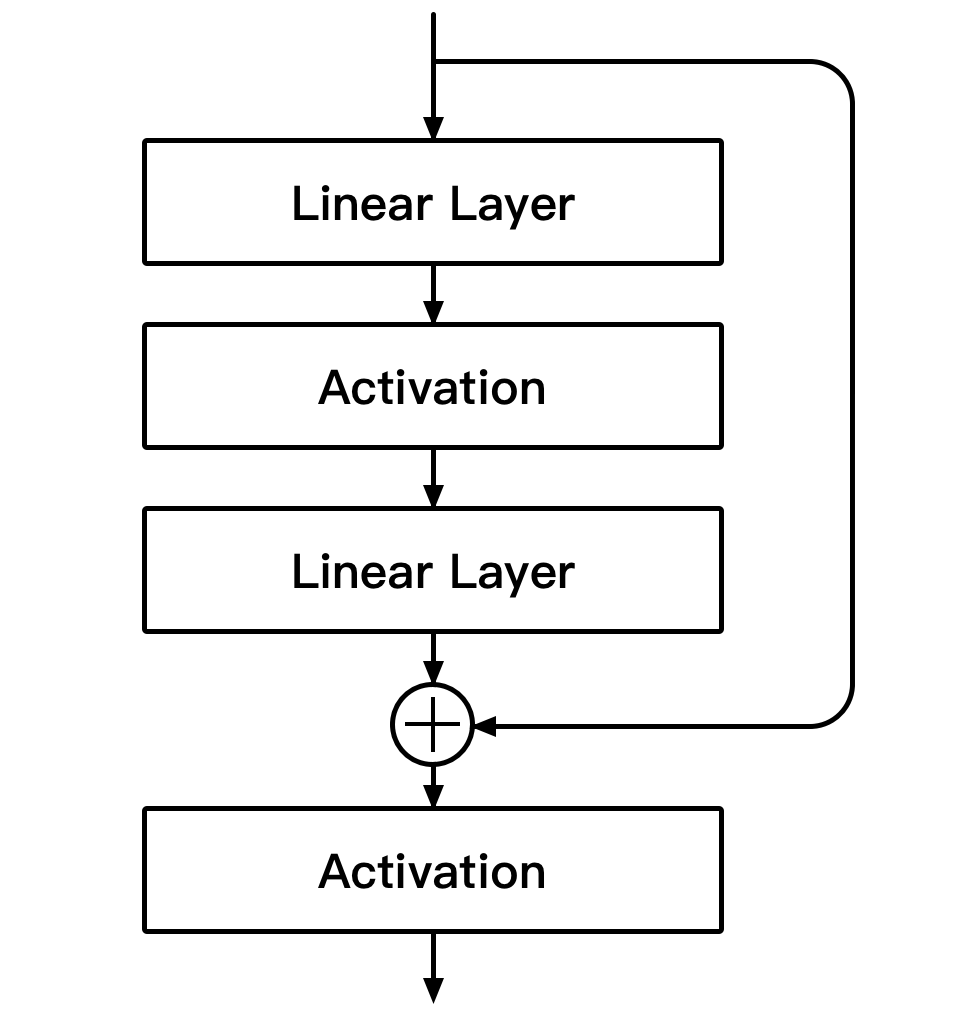}
        \caption{Residual Block}
        \label{fig:residual_block}
    \end{subfigure}

    \caption{Network architecture of residual networks.}
    \label{fig:comparison}
\end{figure}

\begin{table}[h]
    \centering
    \resizebox{\columnwidth}{!}{
    \begin{tabular}{l|c|c|c}
        \toprule
        \textbf{Basic NN} & \textbf{Layers}& \textbf{Input Shape} & \textbf{Output Shape}  \\
        \midrule
        Residual Network& Fully-connected layer& $(B, D_{\text{input}})$& $(B, D_h)$\\
 & Activation& $(B, D_h)$&$(B, D_h)$\\
 & Residual Blocks& $(B, D_h)$&$(B, D_h)$\\
 & Fully-connected layer& $(B, D_h)$&$(B, D_\text{output})$\\
 \midrule
        MLP& Fully-connected layer& $(B, D_{\text{input}})$& $(B, D_h)$\\
 & Activation& $(B, D_h)$&$(B, D_h)$\\
 & Hidden Fully-connected Blocks& $(B, D_h)$&$(B, D_h)$\\
 & Fully-connected layer& $(B, D_h)$&$(B, D_\text{output})$\\
 \bottomrule
    \end{tabular}
    }
    \caption{Basic NNs}
    \label{tab:Basic NNs}
\end{table}

\begin{table}[h]
    \centering
    \resizebox{\columnwidth}{!}{
    \begin{tabular}{l|c|c|c}
        \toprule
        \textbf{Module} & \textbf{Basic NN}& \textbf{Input Shape} & \textbf{Output Shape}  \\
        \midrule
        Position Encoder (M)& Residual Network& $(B, D_q)$& $(B, D_v\times(D_v+1)/2)$ \\
 & Rearrange& $(B, D_v\times(D_v+1)/2)$&$(B, D_v, D_v)$ \\
        State Encoder (B)& Residual Network& $(B, D_q+D_v)$& $(B, D_v)$ \\
        Action Encoder ($\tau$)& MLP& $(B, D_a)$& $(B, D_v)$ \\
        Corrector & Residual Network& $(B, D_q+D_v+D_a)$& $(B, D_v)$ \\
        \bottomrule
    \end{tabular}
    }
    \caption{MoSim Modules}
    \label{tab:MoSim Modules}
\end{table}

\begin{table}[h]
    \centering
    \resizebox{\columnwidth}{!}{
    \begin{tabular}{l|cc|cc|cc|ccc}
        \toprule
        \textbf{Model Size} & \multicolumn{2}{c}{\makecell{\textbf{Position} \\ \textbf{Encoder}}}& \multicolumn{2}{c}{\makecell{\textbf{State} 
        \\ \textbf{Encoder}}}& \multicolumn{2}{c}{\makecell{\textbf{Action} \\ \textbf{Encoder}}}&\multicolumn{3}{c}{\textbf{Correctors}}\\
 & $N_{B}$& $D_h$& $N_{B}$& $D_h$& $N_{B}$& $D_h$& $N_{C}$ & $N_{B}$& $D_h$\\
        \midrule
        Small& 3&64& 3&64& 1&32 &1& 5&64\\
 Medium& 3&64& 3&64&3& 32&1& 5&128\\
        Large& 3&128& 3&128& 3& 128&3& 5&128\\
        \bottomrule
    \end{tabular}
    }
    \caption{MoSim Parameters}
    \label{tab:MoSim Parameters}
\end{table}

For the position encoder and state encoder in the predictor, as well as the corrector, we use residual networks as shown in Figure \ref{fig:residual_network}, with the residual block design illustrated in Figure \ref{fig:residual_block}. For the action encoder in the predictor, we use a standard MLP composed of multiple linear layers and activation layers. In the position encoder, the output vector of length $n(n+1)/2$ (where n is the dimension of velocity) from the residual network is rearranged into an $n\times n$ lower triangular matrix $L$, and the symmetric positive definite matrix $M$ is then obtained by computing $LL^T$.

The number of correctors can be adjusted based on the complexity of the task. We use one corrector for most tasks, while two correctors are used for the humanoid task.

The specific architectural parameters of the models are detailed in Tables \ref{tab:Basic NNs}, \ref{tab:MoSim Modules}, and \ref{tab:MoSim Parameters}, in which $B$ refers to batch size, $D_{\text{input}},\ D_{\text {output}},\ D_h,\ D_q,\ D_v,\ D_a$ respectively refer to dimension of input, output, latent variable, position, velocity, action.

\section{Experimental Details for Latent Evaluation}
\label{Experimental Details}
In subsection \ref{subsec:latent}, we compared the predictive capabilities in the latent space of MoSim and TD-MPC2 using expert data, as described in the main text. For the comparison on random data in table\ref{tab:vstdmpc2}, we still used the provided TD-MPC2 checkpoints, since training TD-MPC2 on random data leads to a meaningless latent space.

In this experiment, 1 step for MoSim is equivalent to \texttt{control step * action repeat}. Here, the control step refers to the duration of a single action in terms of physics timesteps in the DM Control environment, while the action repeat follows the default TD-MPC setting of 2. For example, in the Humanoid environment (\texttt{control step = 5}, \texttt{action repeat = 2}), 1 step actually requires us to recursively predict \texttt{2 * 5 = 10} physical steps, for experience evaluation, we set action repeat = 1, control step = 1.

\section{Prediction Horizon}
\label{Prediction Horizon}
To more directly demonstrate the improvement in MoSim's prediction capability, Table \ref{tab:predction_horizen} uses the MSE loss of DreamerV3 at a 16-step prediction horizon as a reference. It presents the prediction horizon at which MoSim achieves the same loss, providing a more intuitive measure of MoSim's predictive performance.

\begin{table}
\centering
\resizebox{\columnwidth}{!}{
\begin{tabular}{ l  c  c  c  c  c  c  c }
\toprule
Prediction Horizon & Cheetah & Reacher & Acrobot & Panda & Hopper & Humanoid & Go2 \\
\midrule
DreamerV3 &  16&  16&  16&  16&  16&  16&  16\\
\midrule
MoSim &  60&  $>$1000&  99&  $>$1000&  51&  42&  200\\
\bottomrule

\end{tabular}
}
\caption{Prediciton Horizon}
\label{tab:predction_horizen}
\end{table}

\section{Zero-Shot Learning Requirements}
\label{Zero-Shot Learning}

In subsection\ref{subsec:zero_shot}, we explored the minimal step limitation required to achieve zero-shot learning. We additionally provide the requirements for two more tasks. The results are shown in Table \ref{tab:my_label1}.

\begin{table}
    \centering
    \begin{tabular}{ccc}
    \toprule
         Task Name &  Humanoid-Walk& Hopper-Hop\\
         \midrule
         Horizon&  1000& 1000\\
        \bottomrule
    \end{tabular}
    \caption{The minimum step limitation required for zero-shot learning in two additional tasks.(1000 is typically the default step limit in reinforcement learning simulation environments.)}
    \label{tab:my_label1}
\end{table}

\section{MyoSuite Dataset Prediction Results}
We also conducted predictions on the MyoSuite dataset. To satisfy MoSim's assumption of the Markov property, instead of directly using the muscle-tendon model in MyoSuite (where actions in this case form a second-order dynamic system, causing the overall system to violate the Markov property), we directly used the motor-tendon model. Specifically, we obtained the force on each tendon and learned its effect on joint states. 

\section{Experimental Settings for Residual Flow Penalty}
\label{appendix:residual_flow}
To ensure the penalty term is on the same scale as the original reward, we use the log probability density of the current state under the flow-based model as the penalty term and add it directly to the original reward. To normalize the penalty term within the range of \([-1, 1]\), we apply a sigmoid function with a custom inflection point and then subtract 1. This transformation ensures that the penalty term is appropriately scaled for reinforcement learning.

Thus, the final reward function is given by:
\begin{equation}
    R = R_{\text{original}} + R_{\text{penalty}}
\end{equation}
where the penalty term \( R_{\text{penalty}} \) is defined as:
\begin{equation}
    R_{\text{penalty}} =\sigma\left(\frac{\log P(\mathbf{s}) - \tau}{\alpha} \right) - 1 
\end{equation}
where \( \log P(\mathbf{s}) \) is obtained using the normalizing flow model as:
\begin{equation}
    \log P(\mathbf{s}) = \log P_{\text{base}}(x) + \log |\det J|
\end{equation}
with:
\begin{equation}
    x, \log |\det J| = f^{-1}(\mathbf{s})
\end{equation}
where:
\begin{itemize}
    \item \( f^{-1} \) is the inverse transformation of the normalizing flow model,
    \item \( P_{\text{base}}(x) \) is the probability density function of the base distribution (e.g., Gaussian),
    \item \( J \) is the Jacobian matrix of the transformation,
    \item \( \sigma(x) = \frac{1}{1 + e^{-x}} \) is the sigmoid function,
    \item \( \tau \) is the custom inflection point,
    \item \( \alpha \) is a scaling factor.
\end{itemize}

\begin{figure}[H]
    \centering
    \includegraphics[width=\linewidth]{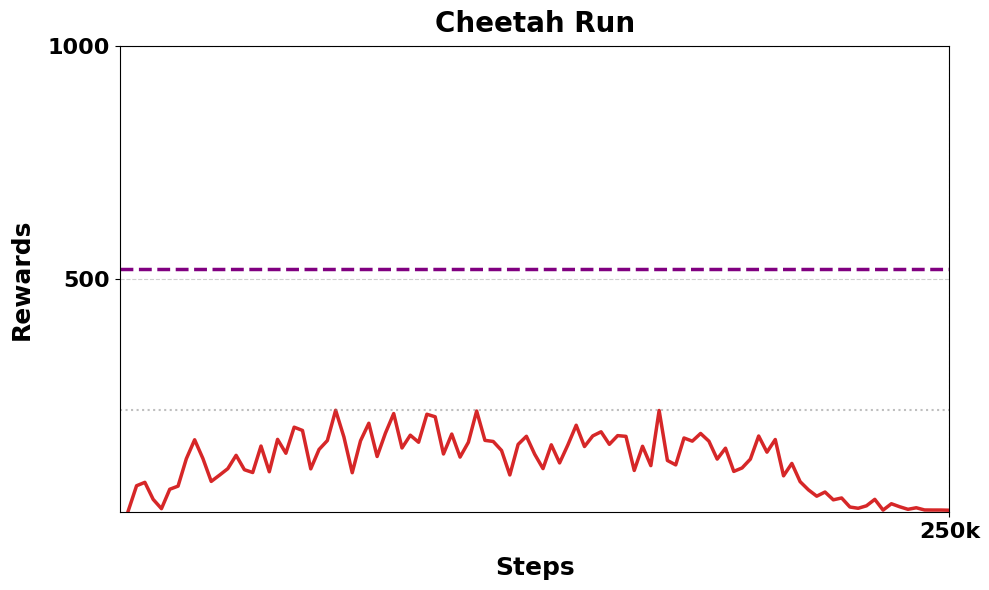}
    \caption{Offline cheetah run.}
    \label{fig:offline}
\end{figure}

\end{document}